\documentclass{ecai} 

\usepackage{latexsym}
\usepackage{amssymb}
\usepackage{amsmath}
\usepackage{amsthm}
\usepackage{booktabs}
\usepackage{enumitem}
\usepackage{graphicx}
\usepackage{color}
\usepackage{multirow}
\usepackage{float}
\usepackage{tikz}
\usepackage{svg}

\usepackage{dsfont}
\usepackage{xcolor}
\usepackage[normalem]{ulem} 

\usepackage{pifont}
\newcommand{\etal}{\textit{et\hspace{0.3em}al.\hspace{0.3em}}}
\newcommand{\quo}[1]{``#1''}
\newcommand*\circled[1]{\tikz[baseline=(char.base)]{
            \node[shape=circle,draw,inner sep=1pt] (char) {#1};}}

\usepackage{algorithm}
\usepackage{algpseudocode}

\newcommand{\BibTeX}{B\kern-.05em{\sc i\kern-.025em b}\kern-.08em\TeX}

\begin{document}


\begin{frontmatter}


\paperid{8733} 


\title{Multi-Attribute Bias Mitigation\\
via Representation Learning}

\author[]{\fnms{Rajeev Ranjan}~\snm{Dwivedi}\orcid{0000-0002-2127-564X}\thanks{Corresponding Author. Email: rajeev22@iiserb.ac.in.}\footnote{Equal contribution.}}
\author[]{\fnms{Ankur}~\snm{Kumar}\footnotemark}
\author[]{\fnms{Vinod}~\snm{K Kurmi}\orcid{0000-0003-0332-5647}}

\address[]{Indian Institute of Science Education and Research (IISER) Bhopal}

\begin{abstract}
Real-world images frequently exhibit multiple overlapping biases, including textures, watermarks, gendered makeup, scene-object pairings, etc. These biases collectively impair the performance of modern vision models, undermining both their robustness and fairness. Addressing these biases individually proves inadequate, as mitigating one bias often permits or intensifies others.  We tackle this multi-bias problem with Generalized Multi-Bias Mitigation (GMBM), a lean two-stage framework that needs group labels only while training and minimizes bias at test time. First, Adaptive Bias-Integrated Learning (ABIL) deliberately identifies the influence of known shortcuts by training encoders for each attribute and integrating them with the main backbone, compelling the classifier to explicitly recognize these biases. Then Gradient-Suppression Fine-Tuning prunes those very bias directions from the backbone’s gradients, leaving a single compact network that ignores all the shortcuts it just learned to recognize. Moreover we find that existing bias metrics break under subgroup imbalance and train–test distribution shifts, so we introduce Scaled Bias Amplification (SBA): a test-time measure that disentangles model-induced bias amplification from distributional differences. We validate GMBM on FB-CMNIST, CelebA, and COCO, where we boost worst-group accuracy, halve multi-attribute bias amplification, and set a new low in SBA—even as bias complexity and distribution shifts intensify—making GMBM the first practical, end-to-end multi-bias solution for visual recognition. Project page: \textcolor{blue}{\url{https://visdomlab.github.io/GMBM/}}

\end{abstract}

\end{frontmatter}


\section{Introduction}
\label{sec:Intro}

In recent years, the remarkable success of machine learning models in image classification has been tempered by growing evidence of their vulnerability to biases in training data. Typically, these biases take the form of shortcuts—spurious correlations or unintended cues that models exploit to boost average performance at the expense of reliability and fairness \cite{geirhos2020NatMachIntellShortcut,torralba_unbiased_look_at_dataset_bias_cvpr_11}. For instance, classifiers might learn to associate water backgrounds with boats \cite{singla2022Int.Conf.Learn.Represent.Salient}, face recognition systems can amplify gender and skin‐tone biases present in their training sets \cite{buolamwini_gender_shades_facct_18}, and in medical imaging, COVID‑19 diagnostic models have been shown to rely on dataset‑specific artifacts—such as hospital tags or watermarks—leading to performance degradation when these cues are absent \cite{degrave2021NatMachIntellAI}.

To counter these issues, a range of bias‑mitigation approaches have been developed. Most of these techniques operate under the assumption of a single, known spurious cue and often demand privileged group annotations during training or validation \cite{sagawa2020Int.Conf.Learn.Represent.Distributionally,nam2020Adv.NeuralInf.Process.Syst.Learning}. However, recent studies demonstrate that this simplification fails to capture the complexity of real‑world datasets—such as ImageNet and popular facial attribute benchmarks—which typically harbor multiple simultaneously; unknown biases like watermarks, textures, and latent correlations. As a result, models trained under these conditions may exploit various spurious features, leading to unpredictable failures when any one of these cues is removed or altered. Although in‑processing strategies like distributionally robust optimization (DRO) and last‑layer retraining can improve worst‑group performance for a solitary bias, they falter in multi‑bias scenarios \cite{sagawa2020Int.Conf.Learn.Represent.Distributionally,kirichenko_last_layer_retraining_bias_mitigation_iclr_23}. Likewise, unsupervised methods that leverage training dynamics can uncover a dominant bias direction but lack the capacity to disentangle more than one bias simultaneously~\cite{jain2022distilling,tsirigotis_no_labels_ssl_bias_mitigation_neurips_23}. 

In addition to this, real‑world datasets often contain multiple overlapping biases that jointly erode model robustness~\cite{lang2021IEEECVFInt.Conf.Comput.Vis.ICCVExplaining,eyuboglu_domino_bias_discovery_iclr_23}. The problem becomes even more challenging when the efforts to suppress one shortcut shifts the reliance onto alternative spurious cues and even amplify it, giving rise to a “Whac‑A‑Mole” dilemma in bias mitigation as stated by Li \etal~\cite{li2023whac}.

\begin{figure}[t]
    \centering
    \vspace{1em}
    \includegraphics[width=1.0\linewidth]{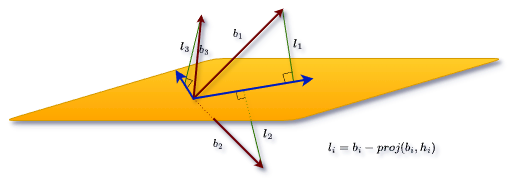}
    \vspace{-2em}
    \caption{Gradient suppression fine tuning}
    \label{fig:teaser}
\end{figure}

Despite practical significance, effective strategies for mitigating multiple interacting biases remain underexplored in the literature. To address multi-bias robustness in vision models, we propose Generalized Multi-Bias Mitigation (GMBM), a hyperparameter-light, two-stage framework that uses group annotations only during training to learn and then suppress multiple spurious-feature representations. Crucially, at inference time, GMBM relies \emph{solely} on the debiased image feature—without bias labels, extra modules, or architectural changes. GMBM performs debiasing in two stages:

\begin{enumerate}
\itemsep 0.4em
  \item[\ding{182}] \textbf{Adaptive Bias–Integrated Learning (ABIL).} 
    For each known bias attribute $j$, we train a encoder whose penultimate-layer output captures that spurious signal. In parallel, the backbone’s penultimate-layer output represents the core image feature. We compute attention weights by applying a softmax over the cosine similarities between the image feature and each bias feature, then fuse the representation by adding the image feature to the weighted sum of bias features. Feeding this fused feature into the classification head forces the model to identify and explicitly discount each spurious channel, disentangling spurious features from task-relevant cues.
  
  \item[\ding{183}] \textbf{Gradient-Suppression Fine-Tuning.} 
    We discard the bias encoders and fine-tune the backbone on clean image feature alone. To eliminate residual bias influence, we first project each bias feature onto the subspace orthogonal to the image feature, obtaining an orthogonal residual, as shown in Fig.~\ref{fig:teaser}. We then use the standard cross-entropy loss with a penalty term that penalizes squared gradient component along each orthogonal residual, scaled by a penalty strength lambda. This enforces invariance to all known biases while preserving legitimate semantic information.
\end{enumerate}

\noindent GMBM is evaluated on both a synthetic dual-shortcut benchmark (with controlled foreground/background color cues) and a real-world CelebA~\cite{liu2015faceattributes} and COCO~\cite{lin2015microsoftcococommonobjects} datasets. At inference time, only the debiased backbone feature is passed to the classifier for efficient deployment without further overhead. Across both settings, GMBM consistently outperforms single-bias baselines and prior multi-bias methods, improving worst-group accuracy by up to 8\% and halving spurious bias amplification.\\[0.5em]
\noindent \textbf{Our key contributions can be summarized as follows:}
\label{subsec:contributions}

(1) We formalize multi-bias mitigation in vision models and critically analyze the limitations inherent in single-bias approaches.
(2) We introduce GMBM, the first end-to-end two-stage framework---comprising ABIL and gradient suppression---that adaptively integrates and subsequently suppresses multiple bias representations.
(3) We create multi-bias evaluation benchmark designed to capture realistic scenarios of intersecting biases.
(4) We provide empirical evidence demonstrating that GMBM establishes a new state-of-the-art in robustness by significantly improving both unbiased and bias-conflicting accuracy while simultaneously decreasing spurious bias amplification.

\section{Related Work}
\label{sec:related_work}

\noindent\textbf{Bias Identification and Discovery.}
A key prerequisite for any mitigation strategy is understanding what biases a model has learned. Recent work has focused on uncovering spurious correlations and underperforming subgroups without explicit bias labels. For example, Eyuboglu \etal \citep{eyuboglu_domino_bias_discovery_iclr_23} leverages cross-modal embeddings and an error‑aware model to pinpoint under‑represented subgroups, while Singla \etal\citep{singla2022Int.Conf.Learn.Represent.Salient} use activation maps to generate Salient ImageNet, a dataset of core vs. spurious feature masks. GSCLIP \citep{zhu2022gsclip} offers a training‑free, dataset‑level shift explanation using CLIP \cite{radford2021learning} embeddings, and generative approaches have been employed to discover unknown biases via latent space manipulation\citep{li2021discover,lang2021explaining}. Jain \etal\citep{jain2022distilling} further distill failure modes with linear probes and CLIP\cite{radford2021learning} captions to explain model errors. While these methods excel at identifying single \quo{shortcuts}, they typically do not scale to settings where multiple, intersecting biases co‑occur. \vspace{1em}

\noindent\textbf{Debiasing in Single-Bias Scenarios.} A large body of work targets spurious correlations between a primary attribute and one secondary (bias) attribute. When bias labels are accessible, several strategies have been developed. These include robust optimization, which re-weights groups based on their performance \citep{sagawa2020Int.Conf.Learn.Represent.Distributionally}; adversarial training, which aims to suppress bias-related signals \citep{dhar2021pass,gong2020jointly,kurmi2022gradient}; and contrastive objectives, which explicitly work to separate examples where the bias aligns with the primary attribute from those where it conflicts \citep{zhang2022correct,tartaglione2021end}. In situations where bias labels are not available, various label-free methods have emerged. These approaches, such as Learned-Mixin (LAD) \citep{LAD}, Environment Inference (EIIL) \citep{creager2021environment}, Just Train Twice (JTT) \citep{JTT}, Learning from Failure (LfF) \citep{NEURIPS2020_eddc3427}, CosfairNet \citep{Dwivedi_2024_BMVC}, and contrastive-based debiasing techniques \citep{zhang2022contrastive}, aim to discover or approximate bias groups by leveraging model-based heuristics. While these methods have proven effective in addressing single spurious correlations, they typically operate under the assumption of \textit{at most one dominant bias}. Consequently, their effectiveness is limited when multiple independent biases interact within the data \citep{lang2021explaining}.

\vspace{1em} \noindent\textbf{Multi-Attribute Bias Mitigation via Representation Learning.}
Methods specifically designed to address the challenge of handling simultaneous, unknown biases have only recently begun to emerge. For example, Li \etal\citep{li2022discover} proposes an iterative approach that assigns pseudo-labels to discover multiple biases and subsequently trains deconfounded models. Similarly, Whac-A-Mole \citep{li2023whac} employs targeted augmentations to simulate a variety of bias types. However, both of these methods are primarily tailored to synthetic image benchmarks and rely on hand-crafted bias generators. In contrast, representation learning techniques offer a promising avenue for addressing an arbitrary number and variety of biases in natural settings. These techniques include invariant feature extraction, information-theoretic regularizers, and multi-view contrastive learning.

In this work, we address these challenges and introduce a unified representation learning-based framework that aims to mitigate multiple overlapping biases by a two-stage network. We present our novel method in the following section~\ref{sec:methodology}.

\section{Methodology}
\label{sec:methodology}

\textbf{Problem Formulation:} We consider an \(N\)-way classification dataset
\[
\mathcal{D} \;=\;\bigl\{\bigl(x^{(i)},\,y^{(i)},\,b^{(i)}_1,\dots,b^{(i)}_k\bigr)\bigr\}_{i=1}^n,
\]
where \(n\) is the total number of samples in the dataset, each input \(x^{(i)}\) carries a ground-truth label \(y^{(i)}\in\{1,\dots,N\}\), also \(b^{(i)}_1,\dots,b^{(i)}_k\) are \(k\) known bias attributes. Let $h^{(i)} \;=\; f_{\mathrm{pen}}(x^{(i)}) \;\in\;\mathbb{R}^d$ denote the penultimate (“pen”) representation extracted by the backbone. We seek a classifier
\[
g:\mathbb{R}^d\;\to\;\{1,\dots,N\}
\]
that predicts \(y^{(i)}\) accurately without exploiting any spurious attribute \(b_j\). Concretely, we call \(b_j\) \emph{spurious} if $H\bigl(Y \mid B_j\bigr)\approx 0,$ where $$H\bigl(Y \mid B_j\bigr) = -\mathbb{E}_{B_j}\Bigl[\sum_{y=1}^N P\bigl(Y=y\mid B_j\bigr)\,\log P\bigl(Y=y\mid B_j\bigr)\Bigr]$$ is the Shannon conditional entropy of the random label \(Y\) given attribute \(B_j\). In settings with multiple such biases, our goal is to ensure that the model’s decision remains correct even when each \(b_j\) is removed or contradicted, thus guaranteeing robustness across all known bias dimensions.

Building on this formulation, we now introduce our two-stage debiasing framework. First, in Section \ref{subsec:ABIL} we describe \emph{Adaptive Bias–Integrated Learning (ABIL)}, which exposes and weights each known bias cue via a soft‐attention mechanism to challenge the model to discount spurious features. Then, in Section \ref{subsec:ITGO}, we detail \emph{Inference‐Time Gradient Orthogonalization}, which fine‐tunes the backbone to enforce invariance to any residual bias directions.

\subsection{Adaptive Bias–Integrated Learning (ABIL)}
\label{subsec:ABIL}
Our goal in ABIL is to \emph{isolate} and then \emph{attenuate} spurious bias cues, while preserving and emphasizing task-relevant information.  We achieve this via a two-stream architecture and a dynamic fusion mechanism, justified as follows:

\vspace{0.5em} \noindent (1) \uline{Bias Encoders.}  
We allocate an encoder per known bias attribute $j$, trained to predict $b^{(i)}_j$ from $X_i$ denoted as $f^j_{\mathrm{pen}}$.  By dedicating separate parameters to each bias, we encourage the network to carve out \emph{distinct} subspaces in $\mathbb{R}^d$ that specialize in capturing that particular spurious signal.  This explicit disentanglement simplifies later suppression.

\vspace{0.5em} \noindent (2) \uline{Penultimate-Layer Representations:} 
Both the main backbone and each bias encoder output features from their penultimate (pen) layer:
$$h^{(i)} = f_{\mathrm{pen}}\bigl(x^{(i)}\bigr),\quad
b_j^{(i)} = f^j_{\mathrm{pen}}\bigl(x^{(i)}\bigr),$$

We use the penultimate layer activations, as they capture a rich, high-dimensional embedding of the input abstractions enough to encode semantic concepts \cite{raghu2017svcca, zhao-etal-2024-layer, stanley2025and, hosseini2023large}, not yet collapsed into class scores.  This makes $h_i$ and each $b_i^j$ suitable for measuring alignment and for controlled fusion, without the interference of the final decision boundary.

\vspace{0.5em} \noindent (3) \uline{Soft-Attention over Bias Channels.}  
Instead of naively concatenating or summing all $b_i^j$, we compute:
$$\alpha_j^{(i)}  = \frac{\exp\left(\cos(h_i, b_i^{j})\right)}{\sum_{m=1}^k \exp\left(\cos(h_i, b_i^{m})\right)},  \quad\quad \cos(u,v) = \frac{u\!\cdot\!v}{\|u\|\|v\|}$$

This ensures that bias encoders whose representations align more strongly with the current image feature $h_i$ receive higher weight, reflecting which spurious cues are \emph{most likely} to influence the backbone. Furthermore, the use of softmax ensures that the attention mechanism remains fully differentiable, which enables a co-adaptive learning process during training. In this process, the backbone learns to produce features that \textit{de-emphasize} subspaces that are subsequently down-weighted. Additionally, employing softmax has two key benefits: it prevents unbounded amplification of bias vectors and yields an interpretable probability distribution over spurious factors.

\vspace{0.5em} \noindent (4) \uline{Bias-Modulated Fusion.}  
We form the composite feature

$$h'_i = h_i + \sum_{j=1}^k \alpha_j^{(i)} \,b_i^j \quad  \xrightarrow \quad \mathcal{L}_{\mathrm{ABIL}} =  \mathrm{CE}\bigl(g(h_i'),\,y_i\bigr)$$
via a residual additive connection. In training we then optimize the standard \emph{cross‐entropy} (CE) loss over our composite features. The backbone’s original feature $h_i$ remains intact, while bias vectors act as \emph{perturbations} highlighting spurious directions. By presenting classifier with both clean and bias-accentuated signals, the network naturally learns to reward reliance on $h_i$’s invariant components and penalize shortcuts through $b_i^j$.

\vspace{0.5em} \noindent (5) \uline{Training with $h'_i$.}  
The fused feature $h'_i$ is fed into the final classification head during training.  This is done in an adversarial framing where the model must solve the task \textit{in the presence of explicit bias cues}. Consequently, the model internalizes robust features in $h_i$ that remain predictive even when spurious channels are subsequently suppressed (in our inference-time orthogonalization).

Together, with this algorithmic setup, ABIL does not merely hide
bias information, but systematically \emph{identifies}, \emph{weights}, and then \emph{challenges} it leading to a backbone representation that cleanly separates task-relevant structure from known spurious factors.

\subsection{Gradient-Suppression Fine-Tuning}
\label{subsec:ITGO}

After ABIL has equipped the backbone with bias‐aware features, we perform a brief fine‐tuning step to \emph{guarantee} that no residual spurious components influence the final prediction. We reuse the bias encoders trained in ABIL to extract $b_i^j$ alongside the backbone feature $h_i$. The residual vectors \begin{equation}
l_i^j = b_i^j - \frac{h_i^\top b_i^j}{\|h_i\|^2} h_i 
\label{eq:1}
\end{equation} capture the pure bias directions that ABIL exposed. By penalizing $$\sum_{j=1}^k (\nabla_{h_i} L_{ce} \cdot l_i^j)^2$$ we suppress any gradient component that would steer $h_i$ back into bias subspaces, thereby enforcing \emph{provable invariance} to all known spurious channels.
\begin{figure*}[ht]
    \centering
    \includegraphics[width=0.7\textwidth]{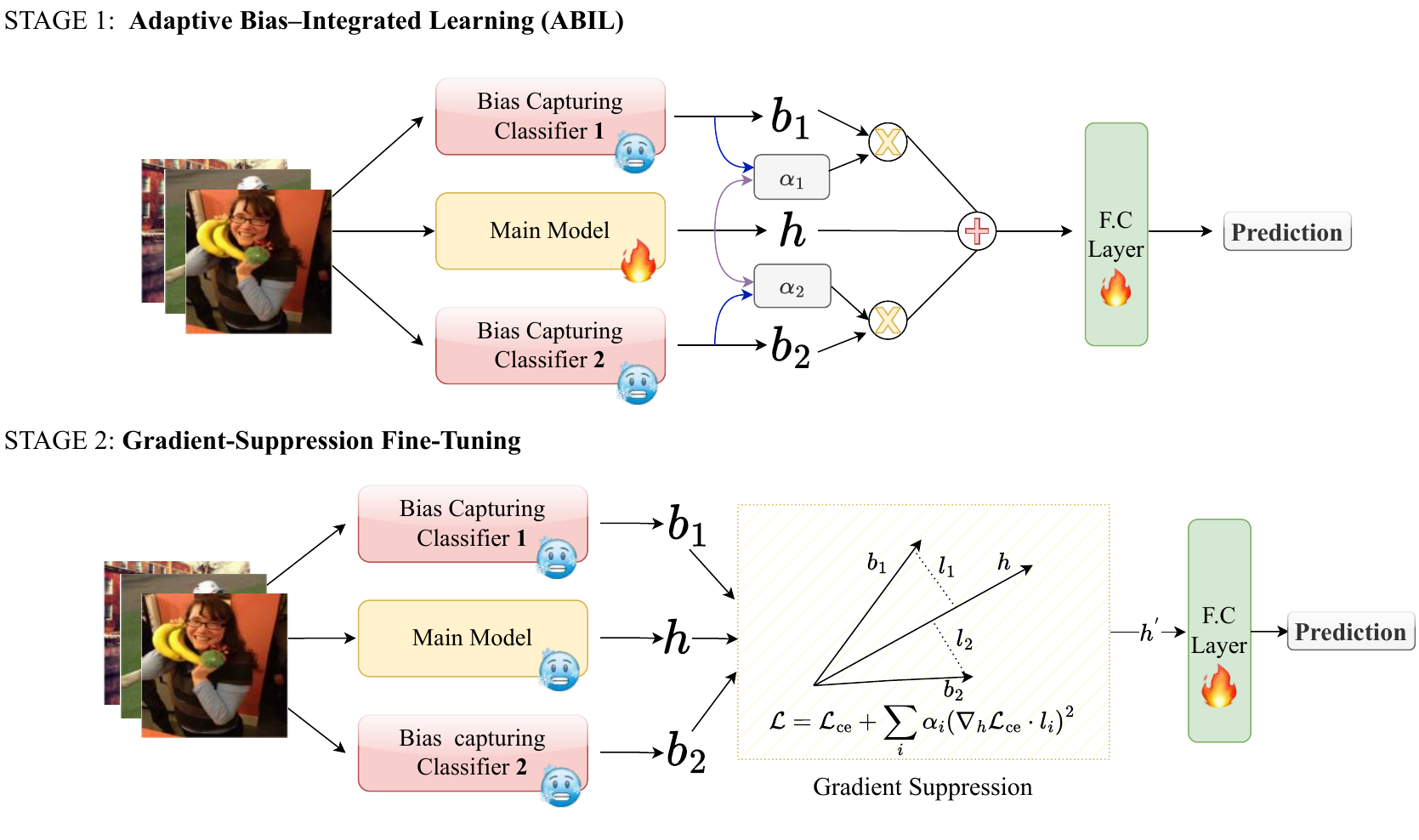}
    \caption{\textbf{Overview of the Generalized Multi-Bias Mitigation (GMBM) Framework.}  Stage 1 (ABIL)—Multiple bias encoders are trained alongside the main model to explicitly capture known spurious features. Their outputs are integrated with the backbone feature, forcing the classifier to learn to discount bias-aligned directions. Stage 2 (GSFT)—after discarding the bias encoders, the model is fine-tuned with a gradient-penalty term that suppresses residual alignment with bias directions. This results in a compact, bias-invariant backbone used at inference time.}

    \label{fig:main}
\end{figure*}

\noindent\textbf{Overall Objective.}  
Our framework consists of two sequential stages—adaptive bias integration followed by gradient‐based fine‐tuning (Fig.~\ref{fig:main}).

\medskip

\noindent\textbf{Stage \circled{1}: Adaptive Bias Integration.}  
During training, each image feature is augmented by a weighted sum of bias‐specific features. We compute attention weights by comparing the image feature to each bias feature using cosine similarity, then normalize these scores via softmax. The resulting fused representation is optimized with standard cross‐entropy loss against the true labels.

\medskip

\noindent\textbf{Stage \circled{2}: Gradient‐Based Fine‐Tuning.}  
In the second phase, we drop all bias encoders and fine‐tune using only the original image feature. To eliminate any remaining bias influence, each bias feature is projected onto the space orthogonal to the image feature, and we add a regularization term that penalizes any gradient component aligned with these projections. A fixed penalty weight of 0.01 ensures a balance between bias suppression and classification accuracy, yielding a single, robust model.

\medskip

\noindent Having described our two‐stage debiasing, we now turn to how we choose and control bias attributes in our evaluation. In Section \ref{subsec:dataset} we summarize three benchmark datasets and specific bias attributes used and give the detailed algorithm \ref{alg:gmbm} for our method.

\begin{algorithm}[ht]
\small
\caption{Generalized Multi‐Bias Mitigation (GMBM)}
\label{alg:gmbm}
\begin{algorithmic}[1]
\Require Dataset $\mathcal D=\{(x,y,b_1,\ldots,b_k)\}$;
         backbone $f$ with penultimate hook $f_{\text{pen}}$;
         classifier $g$;
         bias encoders $f_{\text{pen}}^{1..k}$;
         bias‐classifier heads $\hat b_{1..k}$;
         weights $\beta,\lambda_{\text{supp}}$
\Ensure Debiased backbone $f_{\text{debiased}}$
\Statex \textbf{Stage 1: Adaptive Bias‐Integrated Learning (ABIL)}
\ForAll{minibatch $(X,Y,B_1,\ldots,B_k)$}
    \State $H \gets f_{\text{pen}}(X)$
    \For{$j=1$ \textbf{to} $k$} \Comment{bias representations}
        \State $B_j \gets f_{\text{pen}}^{j}(X)$
    \EndFor
    \State $\alpha_j \gets \operatorname{softmax}\!\bigl(\cos(H,B_j)\bigr)$
    \State $H' \gets H + \sum_{j=1}^{k}\alpha_j\,B_j$
    \State $\mathcal L_{\text{main}} \gets \operatorname{CE}\bigl(g(H'),Y\bigr)$
    \State $\mathcal L_{\text{bias}} \gets \sum_{j=1}^{k}\operatorname{CE}\bigl(\hat b_j(B_j),B_j\bigr)$
    \State Update $f,g,f_{\text{pen}}^{j},\hat b_j$ w.r.t.\ $\mathcal L_{\text{main}}+\beta\,\mathcal L_{\text{bias}}$
\EndFor
\State Freeze $f_{\text{pen}}^{1..k}$; \text{Discard} $\hat b_{1..k}$
\Statex \textbf{Stage 2: Gradient-Suppression Fine-Tuning}
\ForAll{minibatch $(X,Y)$}
    \State $H \gets f_{\text{pen}}(X)$
    \For{$j=1$ \textbf{to} $k$}
        \State $B_j \gets f_{\text{pen}}^{j}(X)$
        \State $L_j \gets B_j \;-\;\dfrac{\langle H,B_j\rangle}{\|H\|^{2}}\,H$
    \EndFor
    \State $\mathcal L_{\text{ce}} \gets \operatorname{CE}\bigl(g(H),Y\bigr)$
    \State $\mathcal L_{\text{grad}} \gets \sum_{j=1}^{k}\bigl(\nabla_{H}\,\mathcal L_{\text{ce}}\;\cdot\;L_j\bigr)^{2}$
    \State Update $f_{\text{pen}},g$ w.r.t.\ $\mathcal L_{\text{ce}}+\lambda_{\text{supp}}\,\mathcal L_{\text{grad}}$
\EndFor
\Return $f_{\text{debiased}}\gets f$
\end{algorithmic}
\end{algorithm}

\section{Experiments \& Evaluation Metric}
\label{sec:exp_and_metric}

\subsection{Datasets}
\label{subsec:dataset}
We evaluate our method on diverse set of multi-attribute bias datasets, each designed to probe model robustness under complex spurious correlations:
\begin{itemize}
\itemsep 0.4em
    \item \textbf{FB-CMNIST} \cite{sarridis2024badd}: This is  a synthetic extension of the Colored MNIST dataset\cite{bahng2020learning}, where each digit is modified with two spurious biases—background color and foreground (digit) color—enabling the study of models under multiple simultaneous correlations. 
    
    \item \textbf{CelebA} \cite{liu2015faceattributes}: CelebA comprises real-world face images annotated with 40 binary attributes, including gender, hair color, age, and various facial accessories. We employ this dataset to study multi-attribute bias in a binary gender classification setting. Concretely, we take the Male attribute as our target label and select \textit{Wearing\_Lipstick} and \textit{Heavy\_Makeup} as spurious attributes since they are known to correlate strongly with the female gender and can introduce significant bias \cite{lang2021explaining}.
    
    \item \textbf{COCO} \cite{lin2015microsoftcococommonobjects}: We construct a custom dataset based on the COCO2017 dataset \cite{lin2015microsoftcococommonobjects} to study gender and object-based biases following the work in \cite{zhao2021understandingevaluatingracialbiases}. Gender labels are derived from image captions by scanning for gender-indicative keywords. To introduce bias labels, we define two object bias categories using COCO instance annotations. \textbf{Bias Category 1} includes various sports and outdoor objects objects while \textbf{Bias Category 2} includes various indoor and kitchen objects.
\end{itemize}

\noindent \textbf{Implementation Details: }For FB-CMNIST, a simple 7-layer convolutional neural network was used as both the model backbone and the bias-capturing classifier. For the CelebA and COCO datasets, the standard ResNet-18 architecture was employed. The CMNIST model was trained for 80 epochs, followed by 10 epochs of fine-tuning. Similarly, the ResNet-18 models were trained for 20 epochs and fine-tuned for an additional 10 epochs. In both cases, we used an initial learning rate of $10^{-3}$ for the main training phase and $10^{-4}$ for fine-tuning, both steps were employed with the Adam optimizer. A batch size of 128 was used across all experiments. Table \ref{tab:hyperparams} summarizes the hyperparameters used. 

\begin{table}[h]
\centering
\caption{Key hyper-parameters used throughout all experiments.}
\setlength{\tabcolsep}{6pt}
\scalebox{0.9}{\begin{tabular}{@{}lcc@{}}
\toprule
Symbol & Meaning & Default \\ 
\midrule
\(T_1\) & ABIL epochs & $6$ \\
\(T_2\) & Suppression epochs & $3$ \\
\(\beta\) & Bias--loss weight & $0.2$ \\
\(\lambda_{\text{supp}}\) & Gradient penalty & $10^{-2}$ \\
\(d\) & Embedding width & $128$ \\ 
\bottomrule
\end{tabular}}
\label{tab:hyperparams}
\end{table}

\subsection{Evaluation Metrics}
In our problem setting, \quo{multiple bias} attributes may be correlated with one another, jointly influencing the model’s predictions and amplifying bias. Prior work has largely relied on two key metrics—\uline{unbiased precision} and \uline{bias-conflicting accuracy}—to evaluate the effectiveness of bias mitigation algorithms. While useful in simple scenarios dominated by a \quo{single bias}, these metrics fall short in capturing the nuanced effects of multiple, interdependent biases. In such settings, evaluating individual unbiased or bias-conflicting accuracies may overlook how well the model disentangles and mitigates the joint influence of correlated biases.\\[0.25em]

\noindent \textbf{Multi-Attribute Bias Amplification (MABA).}  
Zhao \textit{et al.}~\cite{zhao2023menlaundrymultiattributebias} observe that multiple bias attributes can interact to skew predictions in ways that single‐bias metrics cannot account for.  MABA addresses this by examining every combination of bias attributes $m\in\mathcal{M}$ together with each target label $g\in\mathcal{G}$.  For each pair $(m,g)$, one counts the number of training samples in which $m$ and $g$ co‐occur, denoted $\mathrm{co\text{-}occur}_{\mathrm{train}}(m,g)$, and normalizes over all labels to obtain the training bias \vspace{-0.5em}
\[
\mathcal{B}_{train} = \mathrm{bias}_{\mathrm{train}}(m,g)
= \frac{\mathrm{co\text{-}occur}_{\mathrm{train}}(m,g)}
       {\sum_{g'\in\mathcal{G}}\mathrm{co\text{-}occur}_{\mathrm{train}}(m,g')}.
\]
An analogous procedure on the model’s predicted labels yields the test‐time bias distribution $\mathrm{bias}_{\mathrm{test}}(m,g)$.  To focus on meaningful spurious associations, $\Delta_{gm}$ is defined as:
\[
\Delta_{gm}
= \mathds{1}\{\mathcal{B}_{train}>1/|\mathcal{G}|\}\,\bigl(\mathcal{B}_{test}-\mathcal{B}_{train}\bigr),
\]
$\Delta_{gm}$ thereby ignores the attribute–label pairs whose training frequency does not exceed the uniform prior. The overall amplification is then summarized by the average absolute shift
\[
\mathcal{X} = \frac{1}{|\mathcal{M}|}\sum_{m\in\mathcal{M}}\sum_{g\in\mathcal{G}}|\Delta_{gm}|
\]
and the variance $\mathrm{Var}(\Delta_{gm})$, together forming the Multi\textsubscript{MALS} \cite{zhao2023menlaundrymultiattributebias} metric. A well‐debiased model will produce test‐time bias distributions that closely mirror the training distributions, resulting in low $\mathcal{X}$ (minimal average amplification) and low variance (consistent mitigation across all attribute combinations).  By capturing joint, distributional shifts rather than isolated biases, MABA provides a nuanced measure of how spurious correlations propagate through the model.

\vspace{1em}
\textbf{Problems with MABA:} A key limitation of the MABA metric is that it treats all group-attribute pairs equally, irrespective of their frequency within the training dataset. This becomes problematic when certain combinations are severely underrepresented. In such cases, the bias amplification estimate is unstable and sensitive to noise due to the small sample size, and these rare, practically insignificant combinations can disproportionately influence the metric. This equal weighting scheme results in skewed interpretations, especially in imbalanced datasets. To overcome this limitation, we give two variants of the MABA metric: 
\vspace{1em}

\noindent  \ding{182} \textbf{Min-Support MABA:} Exclude the group-attribute pairs that do not meet a minimum support threshold ($\tau$) in the training data. This ensures that amplification is computed only for statistically reliable combinations, resulting in more robust and interpretable bias estimates.

\[
\Delta_{gm}^{\text{min-supp}} =
\mathds{1}\left\{ \text{co\_occur}_{train}(g, m) > \tau \right\}
\cdot \left( \mathcal{B}_{test}- \mathcal{B}_{train}\right)
\]
\[
\text{Min-Support MABA} = \frac{1}{|\mathcal{M}|} \sum_{g \in \mathcal{G}} \sum_{m \in \mathcal{M}} \left| \Delta_{gm}^{\text{min-supp}} \right|
\]

\vspace{1em}
\noindent \ding{183} \textbf{Weighted MABA:} Here we introduce a frequency-based weighting scheme, where each group-attribute pair is weighted in proportion to its occurrence in the training set. This ensures that more representative groups have greater influence on the final amplification score, leading to a more balanced and realistic measurement of model-induced bias.
\[
\Delta_{gm}^{\text{weighted}} = w_{gm} \cdot \left[ \mathcal{B}_{\text{test}}(g, m) - \mathcal{B}_{\text{train}}(g, m) \right]
\]

$$M_w = \frac{1}{|\mathcal{M}|} \sum_{g, m} |\Delta_{gm}^{\text{weighted}}|, \quad w_{gm} = \frac{c_{gm}}{\sum_{g', m'} c_{g'm'}}$$
where $c_{gm} = \text{co\_occur}_{train}(g, m)$.
\vspace{0.5em}

While the Min-Support MABA variant addresses challenges related to severely under-represented subgroups, and Weighted MABA tackles both under- and over-representation, a critical limitation remains: the original MABA metric and the proposed variants falter in the presence of distribution shift between training and test datasets. If the joint distribution of attributes and groups differs between training and test sets, $\Delta_{gm}$ reflects not just model bias but also dataset shift. The MABA metric may combine distributional shifts between the training and test sets with model-induced bias amplification. When the underlying joint distribution of attributes $m$ and groups $g$ differs between the training set ($P_{\text{train}}(m, g)$) and the test set ground truth ($P_{\text{test, actual}}(m, g)$), the difference $\Delta_{gm} = \text{bias}_{\text{test}}(m, g) - \text{bias}_{\text{train}}(m, g)$ may reflect variations in dataset distributions rather than solely the model's tendency to amplify training biases, diminishing the metric's ability to isolate effects specific to the model.

\subsection{Scaled Bias Amplification (SBA)}

To quantify how much our model amplifies existing group–attribute biases on unseen data (test set), we compute SBA using only test‐set counts.  Let $C^{\mathrm{pred}}_{g,m}$ and $C^{\mathrm{actual}}_{g,m}$ be the number of test instances of group \(g\) with attribute \(m\) predicted by the model and observed in the ground truth, respectively.  We first convert these into proportions:
\[
\text{pred}_{g,m}
=\frac{C^{\mathrm{pred}}_{g,m}}{\sum_{g'}C^{\mathrm{pred}}_{g',m}},
\qquad
\text{act}_{g,m}
=\frac{C^{\mathrm{actual}}_{g,m}}{\sum_{g'}C^{\mathrm{actual}}_{g',m}},
\]
and define the amplification gap $\Delta_{g,m}
=\text{pred}_{g,m}-\text{act}_{g,m}.$ To prevent noisy estimates for rare attributes from dominating the score, we weight each gap by
\[
\omega_{g,m}
=\frac{1}{\sqrt{\sum_{g'}C^{\mathrm{actual}}_{g',m}}\;+\;\epsilon},
\]
where:
\begin{itemize}
\itemsep 0.5em
  \item The \(\sqrt{\cdot}\) yields sublinear scaling so that rarer attributes receive higher weight, but not excessively so—balancing sensitivity to true bias amplification against variance from small‐sample noise.
  \item \(\epsilon>0\) ensures numerical stability (avoiding division by zero) and caps the maximum weight when counts are extremely low.
\end{itemize}

Finally, SBA is the average weighted absolute gap over all groups \(G\) and attributes \(M\):
\[
\mathrm{SBA}
=\frac{1}{|G|\;|M|}\sum_{g\in G}\sum_{m\in M}\omega_{g,m}\,\bigl|\Delta_{g,m}\bigr|.
\]
This formulation yields a single, interpretable scalar: sensitive enough to detect bias amplification yet stable under rare‐group noise.  
\vspace{1em}

\noindent\textbf{Key Benefits of SBA:}  
SBA delivers a more robust and interpretable measure of model‐induced bias amplification by relying exclusively on test‐set comparisons and a subgroup‐aware weighting scheme.

\begin{itemize}
  \item \textbf{Robustness to Distribution Shifts.}  
    SBA uses only test‐set ground truth, avoiding the train‐test co‐occurrence mismatches that destabilize MABA. As the bias ratio $q$ grows from 0.90 to 0.99, SBA for ERM grows steadily (0.265$\rightarrow$0.611$\rightarrow$1.077), whereas MABA’s variance explodes ($>$850 at $q=0.99$). In contrast, SBA for BAdd and GMBM remains low and stable across all $q$ (Table~\ref{tab:maba-fb-cmnist}).

  \item \textbf{Interpretability and Subgroup Sensitivity.}  
    A test‐set scaling factor $\omega_{gm}$ weights each group–label pair by its frequency, preventing majority‐group dominance and ensuring rare, bias‐conflicting instances contribute proportionately to the final score.

  \item \textbf{Stable Variance Profiles.}  
    SBA’s variance across subgroups rises with ERM’s increasing bias (0.233$\rightarrow $1.000 $\rightarrow$2.748) but stays low for BAdd and GMBM ($\leq 0.258$) even under extreme training skew (Table~\ref{tab:sba_cmnsit_var}), confirming its resilience to imbalance.
\end{itemize}

This combination of test‐set centric evaluation and subgroup‐aware weighting makes SBA a more dependable fairness metric for real‐world, imbalanced, or multi‐attribute datasets, overcoming key limitations of prior metrics.

\begin{table}[h]
  \centering
  \caption{Variance of SBA (↓ better) on FB-CMNIST test set  demonstrating GMBM’s stability under increasing spurious correlation.}
  \label{tab:sba_cmnsit_var}
  \begin{tabular}{lccc}
    \toprule
    Model & $q=0.90$ & $q=0.95$ & $q=0.99$ \\ \midrule
    ERM        & 0.233 & 1.000 & 2.748 \\
    BAdd       & 0.013 & 0.048 & \textbf{0.256} \\
    GMBM       & \textbf{0.010} & \textbf{0.042} & 0.258 \\
    \bottomrule
  \end{tabular}
\end{table}

\section{Results}
We evaluate GMBM against multiple baselines on three benchmarks: FB-CMNIST with controlled bias ratios ($q = 0.90, 0.95, 0.99$), CelebA (male vs.\ gender-correlated attributes), and a custom COCO split (sports/outdoor vs.\ kitchen/indoor object biases). We report (i) unbiased and bias-conflicting accuracies, (ii) Multi-Attribute Bias Amplification (MABA) variants, and (iii) the proposed Scaled Bias Amplification (SBA) metric.

\subsection{Unbiased \& Bias-Conflicting Accuracy}

Table \ref{tab:results_CMNIST} compares results of our method on FB-CMNIST dataset with varying bias ratios of 0.90, 0.95, and 0.99. GMBM maintains the highest unbiased accuracy across all bias ratios, achieving 96.1\% ($q=0.90$), 91.5\% ($q=0.95$), and 74.6\% ($q=0.99$). This outperforms the strongest baseline (\textbf{BAdd} \cite{sarridis2024badd}) by +0.5, +2.5, and +5.1\%, demonstrating robustness even under extreme bias ratio (Table~\ref{tab:results_CMNIST}).

On CelebA dataset, for male classification with \emph{Wearing\_Lipstick} and \emph{Heavy\_Makeup} as spurious attributes, GMBM attains the best unbiased accuracies (96.7\% and 95.5\%) and bias-conflicting accuracies (94.5\% and 92.0\%), improving over highest baseline by up to +1.5\% in conflict cases (Table~\ref{tab:celeba-bias}). On the COCO dataset, with sports-object and kitchen-object biases, GMBM achieves unbiased accuracies of 83.78\% and 83.19\%, and bias-conflicting accuracies of 83.85\% and 82.35\%, surpassing all baselines.(Table~\ref{tab:coco-bias}).
\begin{table}[H]
  \centering
  \caption{Unbiased accuracy on the FB-CMNIST test set for various methods under bias ratios $q$, highlighting GMBM’s superior robustness compared to single and multi-bias baselines. (\(\uparrow\) is better)}
  \label{tab:results_CMNIST}
  \scalebox{0.95}{\begin{tabular}{lrrr}
    \toprule
    Method   & $q=0.90$ & $q=0.95$ & $q=0.99$ \\
    \midrule
    Vanilla  & 82.5     & 57.9     & 25.5     \\
    BC‐BB \cite{hong2021bb}   & 80.9     & 66.0     & 40.9     \\
    EnD  \cite{tartaglione2021end}    & 82.5     & 57.5     & 25.7     \\
    FLAC  \cite{sarridis2023flac}   & 84.4     & 63.1     & 32.4     \\
    FairKL \cite{barbano2022fairkl}  & 87.6     & 61.6     & 42.0     \\
    BAdd \cite{sarridis2024badd}    & 95.6     & 89.0     & 69.5     \\
    GMBM     & \textbf{96.1}     & \textbf{91.5}     & \textbf{74.6}     \\
    \bottomrule
  \end{tabular}}
\end{table}
\begin{table}[H]
  \centering
  \caption{Unbiased and bias-conflicting accuracies on a 30\% CelebA test split for gender classification. (\(\uparrow\) is better)}
  \label{tab:celeba-bias}
  \scalebox{0.91}{ \begin{tabular}{lcccc}
    \toprule
    Method & \multicolumn{2}{c}{WearingLipstick} & \multicolumn{2}{c}{HeavyMakeup} \\
    \cmidrule(lr){2-3} \cmidrule(lr){4-5}
           & Unbiased & Bias‐conflicting & Unbiased & Bias‐conflicting \\
    \midrule
    Vanilla & 93.2  & 89.1  & 92.0  & 84.7  \\
    FairKL \cite{barbano2022fairkl}  & 82.7  & 74.7  & 84.4  & 77.9  \\
    BC‐BB \cite{hong2021bb}  & 91.6  & 85.8  & 89.7  & 81.8  \\
    EnD  \cite{tartaglione2021end}    & 95.1  & 91.0  & 92.3  & 85.3  \\
    FLAC \cite{sarridis2023flac}    & 95.4  & 91.6  & 93.2  & 87.2  \\
    BAdd \cite{sarridis2024badd}   & 95.8  & 93.0  & 94.9  & 91.0  \\
    GMBM    & \textbf{96.7}  & \textbf{94.5}  & \textbf{95.5}  & \textbf{92.0}  \\
    \bottomrule
  \end{tabular}}
\end{table}

\begin{table}[H]
  \centering
  \caption{Unbiased and bias-conflicting accuracies on the COCO validation set for gender classification with sports/outdoor vs. kitchen/indoor object biases, showing that GMBM outperforms prior methods on both majority and minority bias-conflicting groups. (\(\uparrow\) is better)}
  \label{tab:coco-bias}
 \scalebox{0.91}{ \begin{tabular}{lcccc}
    \toprule
    Method & \multicolumn{2}{c}{Sports Object} & \multicolumn{2}{c}{Kitchen Object} \\
    \cmidrule(lr){2-3} \cmidrule(lr){4-5}
           & Unbiased & Bias‐conflicting & Unbiased & Bias‐conflicting \\
    \midrule
    Vanilla & 70.81 & 64.61 & 73.20 & 67.36 \\
    FairKL \cite{barbano2022fairkl}  & 76.32 & 67.11 & 74.35 & 76.90 \\
    EnD \cite{tartaglione2021end}    & 77.11 & 70.97 & 82.38 & 77.34 \\
    FLAC \cite{sarridis2023flac}    & 80.02 & 77.31 & 80.22 & 79.95 \\
    BAdd \cite{sarridis2024badd}   & 81.28 & 77.81 & 82.91 & \textbf{83.05} \\
    GMBM    & \textbf{83.78} & \textbf{83.85} & \textbf{83.19} & 83.35 \\
    \bottomrule
  \end{tabular}}
\end{table}

\subsection{Multi-Attribute Bias Amplification (MABA) Variants}
To assess how models amplify existing group–attribute biases, we compute two MABA variants: \emph{Min-Support} and \emph{Weighted} MABA.

On FB-CMNIST, under distribution shift between train/test, MABA metrics exhibit high variance even for debiased models (variance $>850$ at $q=0.99$), highlighting their instability when co-occurrence statistics diverge (Table~\ref{tab:maba-fb-cmnist}). On the CelebA, GMBM achieves the lowest Min-Support MABA mean (0.67) and variance (0.61), improving over ERM (mean 0.74, var 0.85) and BAdd (mean 0.90) (Table~\ref{tab:maba-celeba}). Similarly, on COCO dataset, GMBM again yields the most reliable bias estimates: Min-Support MABA mean 0.73 (vs.\ 8.53 for ERM) and variance 1.66 (vs.\ 96.27), demonstrating consistent bias suppression (Table~\ref{tab:maba-coco}). Note that f\textit{or weighted MABA, the variance is not reported, since group-wise scaling distorts the interpretability of variance.}

\begin{table}[ht]
\centering
\caption{Comparison of Base MABA, Min-Support and Weighted MABA means and variances for ERM, BAdd \cite{sarridis2024badd}, and GMBM on FB-CMNIST across bias ratios $q$, illustrating the high volatility of traditional MABA metrics under distribution shift and GMBM’s relative consistency.}
\label{tab:maba-fb-cmnist}
\scalebox{0.95}{
\begin{tabular}{cllrrr}
\toprule
$q$    & MABA Variant  & Metric    & ERM    & BAdd \cite{sarridis2024badd}  & GMBM   \\
\midrule
\multirow{6}{*}{0.90}
       & \multirow{2}{*}{Base MABA}
                       & Mean     & 14.34 & 16.78 & 16.69 \\
       &                   & Variance & 393.01 & 519.94 & 513.69 \\ 
       & \multirow{2}{*}{Min Support}
                       & Mean      & 14.24  & 16.67  & 16.68  \\
       &                   & Variance  & 393.90 & 513.88 & 513.42 \\ 
       & \multirow{2}{*}{Weighted}
                       & Mean      & 10.98  & 12.69  & 12.69  \\
       &                   & Variance  & -    & -    & -    \\
\midrule
\multirow{6}{*}{0.95}
       & \multirow{2}{*}{Base MABA}
                       & Mean     & 14.41 & 12.41 & 12.38 \\
    &                      & Variance & 384.42 & 381.58 & 388.78 \\
       & \multirow{2}{*}{Min Support}
                       & Mean      & 14.34  & 12.43  & 12.29  \\
       &                   & Variance  & 378.11 & 382.13 & 379.09 \\ 
       & \multirow{2}{*}{Weighted}
                       & Mean      & 10.71  &  7.75  &  7.79  \\
       &                   & Variance  & -    & -    & -    \\
\midrule
\multirow{6}{*}{0.99}
        & \multirow{2}{*}{Base MABA}   
                       & Mean     & 14.54 & 22.82 & 26.66 \\
            &              & Variance & 388.18 & 679.15 & 857.12 \\
       & \multirow{2}{*}{Min Support}
                       & Mean      & 14.69  & 22.82  & 26.64  \\
       &                   & Variance  & 385.48 & 679.72 & 853.92 \\ 
       & \multirow{2}{*}{Weighted}
                       & Mean      & 10.35  & 16.62  & 18.84  \\
       &                   & Variance  & -    & -    & -    \\
\bottomrule
\end{tabular}}
\end{table}

\begin{table}[H]
  \centering
  \caption{Performance of MABA variants on CelebA 30\% test split}
  \label{tab:maba-celeba}
  \scalebox{0.98}{
  \begin{tabular}{llrrr}
    \toprule
    MABA Variant  & Metric    & ERM   & BAdd \cite{sarridis2024badd} & GMBM  \\
    \midrule
    \multirow{2}{*}{Base MABA} 
                    & Mean     & 0.74 & 0.90 & 0.67 \\
                    & Variance & 0.85 & 1.05 & 0.61 \\
    \midrule
    \multirow{2}{*}{Min support} 
                  & Mean      & 0.74  & 0.90  & 0.67  \\
                  & Variance  & 0.85  & 1.05  & 0.61  \\
    \midrule
    \multirow{2}{*}{Weighted} 
                  & Mean      & 0.86  & 1.03  & 0.70  \\
                  & Variance  & -   & -   & -   \\
    \bottomrule
  \end{tabular}}
\end{table}
\vspace{-1em}

\begin{table}[H]
  \centering
  \caption{MABA metrics on COCO‐validation set}
  \label{tab:maba-coco}
  \scalebox{0.95}{\begin{tabular}{llrrr}
    \toprule
    MABA Variant  & Metric    & ERM    & BAdd \cite{sarridis2024badd}  & GMBM   \\
    \midrule
    \multirow{2}{*}{Base MABA} & Mean & 8.549 & 9.46 & 7.88 \\
                             & Variance & 628.50 & 634.83 & 625.08 \\
    \midrule                        
    \multirow{2}{*}{Min support} 
                  & Mean      & 8.53   & 3.53   & 0.73   \\
                  & Variance  & 96.27  & 14.65  & 1.66   \\
    \midrule
    \multirow{2}{*}{Weighted} 
                  & Mean      & 8.84   & 2.60   & 0.35   \\
                  & Variance  & -    & -    & -    \\
    \bottomrule
  \end{tabular}}
\end{table}

Despite these improvements, the high variability of MABA metric under distribution shift motivates our SBA metric.

\subsection{Scaled Bias Amplification (SBA)}
SBA quantifies how much a model's predictions exaggerate the correlation between different groups and attributes when evaluated on a test set. Unlike some other metrics, it is calculated using only the test data, which makes it more reliable when the biases appearing in the training data are different from the test data. It also uses a weighting system to ensure that rare combinations of groups and attributes are not ignored, providing a more balanced view of bias across different subgroups.
\begin{table}[ht]
  \centering
  \caption{SBA scores for ERM, BAdd \cite{sarridis2024badd}, and GMBM across FB-CMNIS, CelebA, and COCO datasets, showcasing GMBM’s effectiveness at maintaining low and stable bias amplification on unseen data. ( $\downarrow$ better )}
  \label{tab:sba}
  \scalebox{0.95}{\begin{tabular}{lcrrr}
    \toprule
    Dataset   & Bias Ratio & ERM  & BAdd \cite{sarridis2024badd} & GMBM \\
    \midrule
    \multirow{3}{*}{CMNIST}
              & 0.90       & 0.26 & 0.05 & \textbf{0.05} \\
              & 0.95       & 0.61 & 0.11 & \textbf{0.10} \\
              & 0.99       & 1.07 & 0.32 & \textbf{0.32} \\
    \midrule
    CelebA    & --         & 0.37 & 1.39 & \textbf{0.35} \\
    \midrule
    COCO      & --         & 0.84 & 0.15 & \textbf{0.10} \\
    \bottomrule
  \end{tabular}}
\end{table}

Under ERM, SBA increases sharply with bias ratio (0.26 $\to$ 0.61 $\to$ 1.07), whereas GMBM remains low and stable (0.05 $\to$ 0.11 $\to$ 0.32), confirming effective mitigation of spurious amplification on the FB-CMNIST dataset (Table~\ref{tab:sba}). On CelebA, GMBM achieves SBA 0.35 vs.\ ERM 0.37 and BAdd 1.39; on COCO, SBA is 0.10 for GMBM vs.\ 0.84 (ERM) and 0.15 (BAdd). Across all settings, GMBM consistently yields the lowest, consistent, and interpretable SBA scores, underscoring its superior ability to prevent bias amplification in unseen data (Table~\ref{tab:sba}).

GMBM not only attains state-of-the-art unbiased and bias-conflicting accuracies across synthetic and real-world benchmarks, but also demonstrably curtails multi-attribute bias amplification—both under traditional MABA variants and our more robust SBA metric. These results validate GMBM’s effectiveness in learning and suppressing multiple spurious features.

\vspace{-0.5em}

\section{Discussion \& Future Work}

Integrating multiple bias representations via attention–weighted fusion steers the backbone toward the most influential shortcuts, driving SBA down to $0.05$ on FB-CMNIST and $0.10$ on COCO while simultaneously boosting bias-conflicting accuracy. Yet GMBM still requires group labels (bias attributes) during training. While this is a reasonable assumption for controlled benchmark datasets, it becomes challenging in many real-world applications where such detailed annotations are often unavailable. A second limitation is its sensitivity to biases that are nearly inseparable from the target label: on CelebA, SBA of $0.35$ suggests that gender remains partly entangled with make-up cues despite mitigation.

Future work can relax these constraints by augmenting ABIL with unsupervised bias discovery. One avenue is to cluster systematic failure modes or latent shortcut directions uncovered by linear probes, echoing the approach of Jain et al. \cite{jain2022distilling}; the discovered prototypes could be fed back into the same gradient-suppression stage to remove unknown biases without extra labels. Complementarily, an online gating mechanism that attenuates any feature whose gradient persistently correlates with emerging shortcut clusters would let GMBM adapt post-deployment. These extensions would broaden applicability while preserving the label-light character of the framework.

\vspace{-0.5em}
\section{Conclusion}

We present GMBM, a two-stage framework that (i) learns to expose and attenuate multiple spurious cues via attention-based bias integration and (ii) enforces invariance through gradient-suppression fine-tuning. In tandem, we introduce the \emph{Scaled Bias Amplification (SBA)}: a test-time metric that quantifies the extent to which a model exaggerates group–attribute correlations under distributional shifts, while normalizing for subgroup over- and under-representation. Our experiments on synthetic and real-world image-classification benchmarks show that GMBM delivers state-of-the-art unbiased and bias-conflicting accuracies while dramatically reducing bias amplification. By tackling multi-attribute spurious correlations in a label-light, inference-efficient way, we advance fairness in vision models and pave the way for adaptive, domain-aware debiasing strategies.

\begin{ack}
Rajeev is supported by the TCS Research fellowship (Cycle 18).
\end{ack}
\vspace{-1em}
\bibliography{mybibfile}

\end{document}